\documentclass[sigconf]{acmart}

\usepackage{todonotes}

\AtBeginDocument{%
  \providecommand\BibTeX{{%
    \normalfont B\kern-0.5em{\scshape i\kern-0.25em b}\kern-0.8em\TeX}}}

\setcopyright{none}
\renewcommand\footnotetextcopyrightpermission[1]{}

\acmConference[WWW '23]{Make sure to enter the correct
  conference title from your rights confirmation emai}{April 03--May 04, 2023}{Texas, USA}
\usepackage{bbding}
\usepackage{multirow}
\usepackage{array}
\usepackage{arydshln}
\usepackage{soul}
\usepackage{xcolor,colortbl}
\usepackage{appendix}
\usepackage{enumitem}
\usepackage{booktabs}
\usepackage{algorithm}
\usepackage{algorithmic}
\definecolor{LightGreen}{rgb}{0.8,1,0.89}
\definecolor{LightRed}{rgb}{1.0,0.8,0.7}
\definecolor{LightCyan}{rgb}{0.88,1,1}
\usepackage{amsmath,amsfonts}

\newcommand{\model}{\texttt{READER}}

\begin{document}
\title{Response-act Guided Reinforced Dialogue Generation \\for Mental Health Counseling}

\author{Aseem Srivastava\textsuperscript{$1$}, Ishan Pandey\textsuperscript{$1$}, Md. Shad Akhtar\textsuperscript{$1$}, Tanmoy Chakraborty\textsuperscript{$2$} }
\affiliation{
\country{
    {
        \textsuperscript{$1$}IIIT-Delhi, India; \textsuperscript{$2$}IIT Delhi, India
    }
}
}

\email{{aseems, ishan20304, shad.akhtar}@iiitd.ac.in; tanchak@iitd.ac.in}

\renewcommand{\shortauthors}{Aseem Srivastava et. al.}

\begin{abstract}

Virtual Mental Health Assistants (VMHAs) have become a prevalent method for receiving mental health counseling in the digital healthcare space. 
% ==
An assistive counseling conversation commences with natural open-ended topics to familiarize the client with the environment and later converges into more fine-grained domain-specific topics. Unlike other conversational systems, which are categorized as open-domain or task-oriented systems, VMHAs possess a hybrid conversational flow.
% ==
These counseling bots need to comprehend various aspects of the conversation, such as dialogue-acts, intents, etc., to engage the client in an effective and appropriate conversation. Although the surge in digital health research highlights applications of many general-purpose response generation systems, they are barely suitable in the mental health domain -- the prime reason is the lack of understanding in the mental health counseling conversation.  Moreover, in general, dialogue-act guided response generators are either limited to a template-based paradigm or lack appropriate semantics in dialogue generation. To this end, we propose \model\ -- a \textbf{RE}sponse-{\bf A}ct guided reinforced \textbf{D}ialogue gen\textbf{ER}ation model for the mental health counseling conversations. \model\ is built on transformer to jointly predict a potential dialogue-act $d_{t+1}$ for the next utterance (\textit{aka} response-act) and to generate an appropriate response ($u_{t+1}$). Through the transformer-reinforcement-learning (TRL) with Proximal Policy Optimization (PPO), we guide the response generator to abide by $d_{t+1}$ and ensure the semantic richness of the responses via BERTScore in our reward computation. We evaluate \model\ on HOPE, a benchmark counseling conversation dataset and observe that it outperforms several baselines across several evaluation metrics -- METEOR, ROUGE, and BERTScore. We also furnish extensive qualitative and quantitative analyses on results, including error analysis, human evaluation, etc.

\end{abstract}

%   <concept>
%       <concept_id>10010147.10010178.10010179.10010182</concept_id>
%       <concept_desc>Computing methodologies~Natural language generation</concept_desc>
%       <concept_significance>500</concept_significance>
%       </concept>
%   <concept>
%       <concept_id>10010147.10010178.10010179.10010181</concept_id>
%       <concept_desc>Computing methodologies~Discourse, dialogue and pragmatics</concept_desc>
%       <concept_significance>500</concept_significance>
%       </concept>
%  </ccs2012>

% \ccsdesc[500]{Computing methodologies~Natural language generation}
% \ccsdesc[500]{Computing methodologies~Discourse, dialogue and pragmatics}

% \keywords{Dialogue system, Response generation, Reinforcement Learning, Mental Health}


\maketitle

\begin{figure}[t]
  \centering
  \includegraphics[width=\columnwidth]{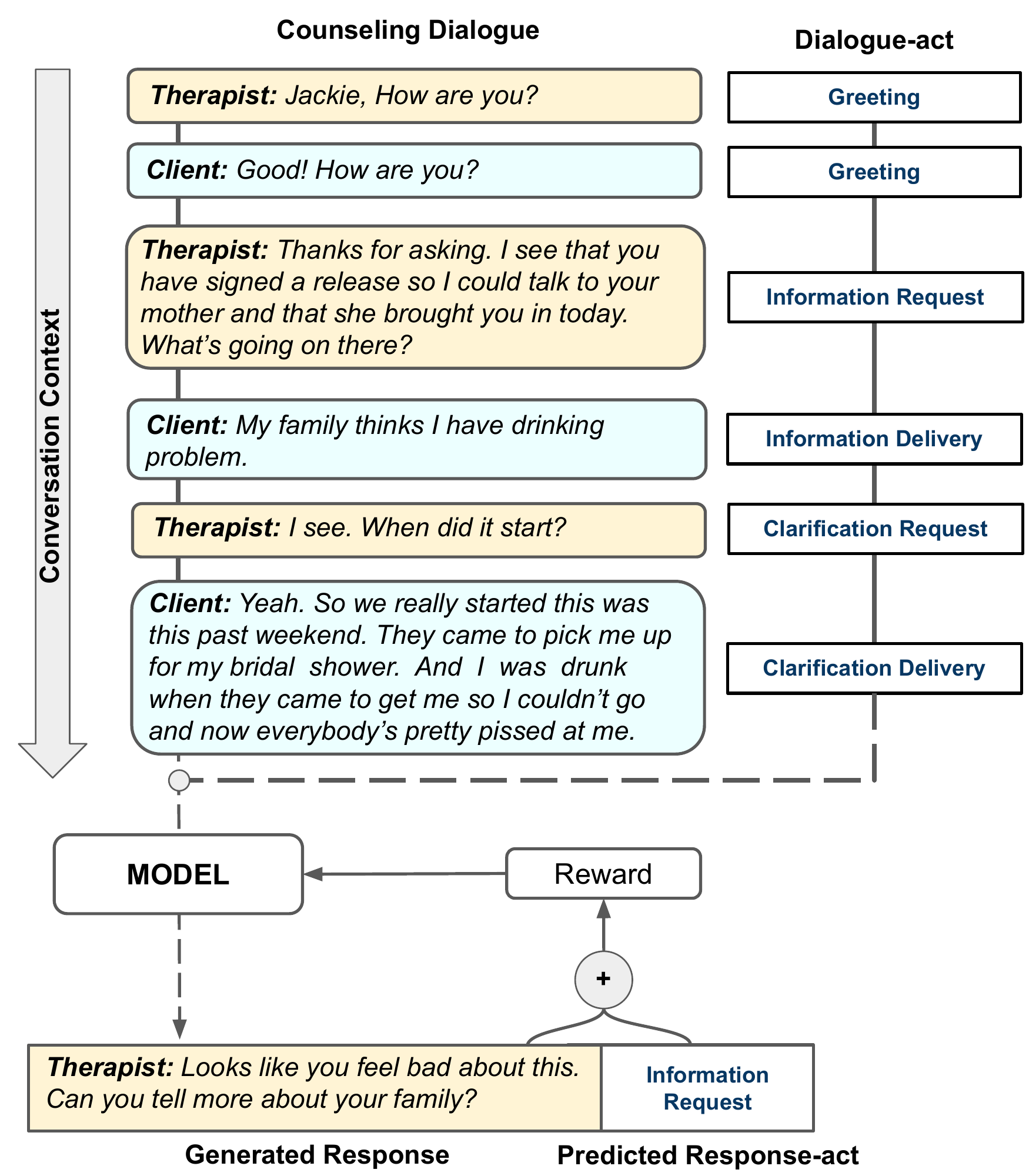}
  \caption{A sample counseling conversation along with associated dialogue-acts. The proposed model -- \model\ takes utterance- and dialogue-act context to predict response-act and subsequently generate a response.}
  \label{fig:taskfigure}
  %\vspace{-5mm}
\end{figure}

\section{Introduction}
\noindent Virtual Mental Health Assistants (VMHAs) are the backbone of the new-age digital healthcare industry. More than 60\% of therapies conducted in the past three years are via virtual assistants. This massive spike in the number of users using VMHAs to gain mental health assistance is due to the ease and safety of access to AI-based therapist-bots \cite{info:doi/10.2196/17828}. 
Numerous potential platforms, \textit{viz.} Weobot\footnote{https://woebothealth.com/}, Wysa\footnote{https://www.wysa.io/}, etc. in the digital health space are developing practical and effective ways for the common public. More popular than ever, VMHAs are now becoming an instant solution to millions of clients struggling with mental health issues\footnote{\url{https://psychnews.psychiatryonline.org/doi/10.1176/appi.pn.2022.05.4.50}}.

 {\bf Limitations of existing methods.} Many such conversational agents fail to (a) understand the directives of the client with whom they are in active conversation, and (b) take the conversation in the required direction\footnote{https://www.bbc.com/news/technology-46507900}. This is resemblant to the fact that even human therapists find it impossible to reply to something they do not understand from the help-seeker. 
Therefore, clients’ directives directly impact the response generation capability. Current open domain conversational systems \textit{viz.} XiaoIce \cite{xiaoice} and GPT-3-based systems \cite{ZHANG2021831} generate semantically and grammatically rich responses.  
However, these open-domain counseling systems lack contextual understanding in the response generation process, which includes being unable to respond with the intended dialogue-act. 
Therefore, modeling this problem using open-domain dialogue systems cannot suffice the task of goal-oriented dialogue systems. To mitigate the issue in the mental healthcare domain,  there is a need to harmonize VMHA's dialogue with individual intentions to be useful for clinical practice. A very generic solution to this is to design a model that gauges the dialogue context and predicts the next dialogue-act (\textit{aka} response-act), which collectively helps generate the next utterance.

 {\bf Our approach.} Our work focuses on response generation by exploiting response-acts. To understand the problem better, Figure \ref{fig:taskfigure} shows an example of a counseling conversation. As we observe, the dialogue contains both therapist's and client's utterances, each possessing a dialogue-act that is critical in maintaining the flow of the conversation. Evidently, the dialogue-acts of the utterances generally form a pattern. For instance, the dialogue-act of the third utterance from the therapist is \textit{information-request}, which in succession is followed by \textit{information-delivery} in the fourth utterance by the client. 
Earlier approaches \cite{noble-maraev-2021-large, chao-etal-2021-improve, Saha2020EmotionAD} exploited dialogue-act and context to build rich representation for several tasks on dialogue system. 
Another work proposes a dialogue management strategy in order to improvise on response generation task exploiting fine-grained belief states \cite{NEURIPS2020_e9462095}. These fine-grained belief states are task-specific, and their proposed system, SimpleTOD, needs relatively more information (specific slots) in order to generate a response. At the same time, exploiting the slot-filling task to generate responses supports the dialogue system in most goal-oriented cases. However, counseling conversations cannot be categorized either into open-ended or goal-oriented dialogue and hence needs a separate focus on the hybrid conversational pattern.
On the other hand, several studies utilize other guiding factors such as keyword, target, etc. for the response generation task \cite{gupta-etal-2022-target, tang-etal-2019-target}. In our work, we predict the dialogue-act of the next utterance \textit{aka} response-act and take advantage of state-of-the-art language models to generate relevant responses. At the same time, contextual information in the conversation plays an essential role in developing a full-fledged conversational system. To this end, we propose a response-act guided dialogue generation model, named, \model. It comprises a foundation language model, on top of which we deploy three unique heads, namely, the response-act head (RAC-Head), the language model head (LM-Head) and the value head (V-Head). These three heads jointly learn to optimize the reinforced loss and primarily perform the response generation task. \model\ learns by optimizing Proximal Policy Optimization (PPO), for which we curate a unique reward function.

 {\bf Evaluation.} 
% \todo{need to write about the evaluation, dataset etc,}
We benchmark \model\ on the HOPE dataset \cite{malhotra2021speaker}, which is a dyadic counseling conversation dataset containing ~$13k$ utterances from therapist and client. We observe that \model\ outperforms several baselines across three relevant quantitative metrics -– METEOR, ROUGE, and BERTScore, with improvements in the range of $0.82 - 11.53\%$. In addition, we also present an extensive qualitative and quantitative analyses of the performance, error analysis and human evaluation. Furthermore, to evaluate the generalizability of the \model, we benchmark it over the Switchboard Dialogue-act corpus (SWDA) \cite{stolcke-etal-2000-dialogue} and obtain better results than baselines by $0.1 - 9.4\%$. 

% \todo{how many baselines, imporvement, etc}

\textbf{Major Contributions.} Below, we summarize the contributions:
\begin{itemize}[leftmargin=*]
    \item We exploit future dialogue-acts (\textit{aka} response-acts) in guiding the response generation model to generate the intended response and maintain the flow of counselling conversation in the mental-health domain. To the best of our knowledge, ours is one of the first attempts that exploits response-acts to generate precise responses in VMHAs or any other dialogue systems.
    \item We propose a novel transformer-reinforcement-learning (TRL) driven response-act guided model, \model\ to generate response in mental health counseling conversations. 
    \item Our evaluation on the HOPE dataset shows significant improvements in the performance of response generation over several competing baselines.  
    % \item We benchmark our proposed model, \model\ on the HOPE Dataset over the metrics ROUGE, BERTScore, and METEOR. We show that our model performs significantly better than the competing baselines.
    We also perform extensive ablation analysis and justify the choice of various components of \model. 
    \item We conduct a through and qualitative human evaluation on the generated responses and establish that the proposed approach is qualitatively efficient as well.
    % . generations from \model\, we show the need of it in real-world applications. This is further supported by the results we obtain from the human evaluation, which shows the efficiency of the proposed method.     
    \item We also show the effective generalizability of \model\ on another dataset, i.e., the Switchboard Dialogue-act dataset. 
    % Exploiting next dialogue-acts (\textit{aka response-acts}), we propose a novel transformer-reinforcement-learning (TRL) driven response-act guided model, \model\ to generate response in mental health counseling conversations.
    % \item We benchmark our proposed model, \model\ on the HOPE Dataset over the metrics ROUGE, BERTScore, and METEOR. We show that our model performs significantly better than the competing baselines.
    % \item Through extensive qualitative analysis on the generations obtained from \model\, we show the need of it in real-world applications. This is further supported by the results we obtain from the human evaluation, which shows the efficiency of the proposed method.     
    % \item  Viewing the sensitivity of the research area, we discuss ethical  considerations. In the last, we demonstrate the generalizability of the \model\ on other datasets.
\end{itemize}

\noindent We have open-sourced the code for \model\ and a sample of the dataset on an anonymous link\footnote{Code and dataset sample: https://bit.ly/3CqpcrK. We commit to publicly release the source codes and datasets upon acceptance of the paper.}.
% There is a huge gap in the treatment that should be available and of the help available at hand, easily and cost-effectively. Even in developed countries, the ratio of psychiatrists, psychologists, psychiatric social workers, and mental health nurses to patients is 1: 10,000. 

\section{Related Work}
To bring more clarity in understanding the role of dialogue-acts, we present relevant studies in two broad areas -- (i) dialogue-act classification, and (ii) dialogue/response generation. We intend to comprehend how dialogue-acts could bring effective innovation in building a conversational system for a dedicated task.

 \textbf{Dialogue-act Classification.}
%First, we understand the roles of dialogue-act in dialogue systems. 
Earlier studies by \citet{budzianowski-etal-2018-multiwoz} and \citet{inproceedings2} employ a sparse representation of each dialogue-act in the form of triple vectors (domain-action-slot); this triple vector is represented as a one-hot encoding. However, acts become very large with the use of such sparse representations. Later, \citet{chen-etal-2019-working} addressed the issue by considering dialogue-act structures. Further, the authors represented dialogue-acts considering the act structures with level-wise vectorization on a one-hot scale where a binary classifier predicts each dimension of vectors. Their methods are further improved in a recent work by \citet{DBLP:journals/corr/abs-1911-08151}. The authors exploited a separate expert decoder for different areas and dialogue-acts to fuse them with a main \textit{chair} decoder.
A recent work applies a fusion approach to fuse their language model with a next utterance generation decoder \cite{inproceedings}. Several other studies use reinforcement learning to generate dialogue responses \cite{DBLP:journals/corr/abs-1909-08593}.

\textbf{Response Generation.}
Studies on dialogue generation \cite{peters-etal-2018-deep,devlin-etal-2019-bert} showed improved performance by leveraging data corpus size, which in turn resulted in learning better context-sensitive features from large language models. \citet{10.5555/3454287.3454804} extended this idea further by deploying models with large parameters. They used a similar idea on XLNet, a generalized autoregressive pretrained model, in order to (i) maximizing the expected likelihood over all permutations of the factorization order allowing learning of bidirectional contexts, and (ii) coping up with the drawbacks of BERT by leveraging the proposed approach's autoregressive formulation. 

Later \citet{radford2019language} explored their hypotheses of the zero-shot learning capacity of large language models as multi-task learners on the task of response generation using GPT-2. The authors also showed an intuitive qualitative analysis of a sample to fetch quality insights. The analysis shows the reflection of coherent responses to prompts. The result presents a better path toward building a response generation system that learns to perform the task from their naturally occurring demonstrations. A recent study on transformer-based models has been fine-tuned for dialogue modeling through various data modification techniques. This includes methods such as adding information about the user's persona, masking, etc. \cite{Wolf2019TransferTransfoAT}. 

\begin{figure*}[ht]
  \centering
  \includegraphics[width=1.0\textwidth]{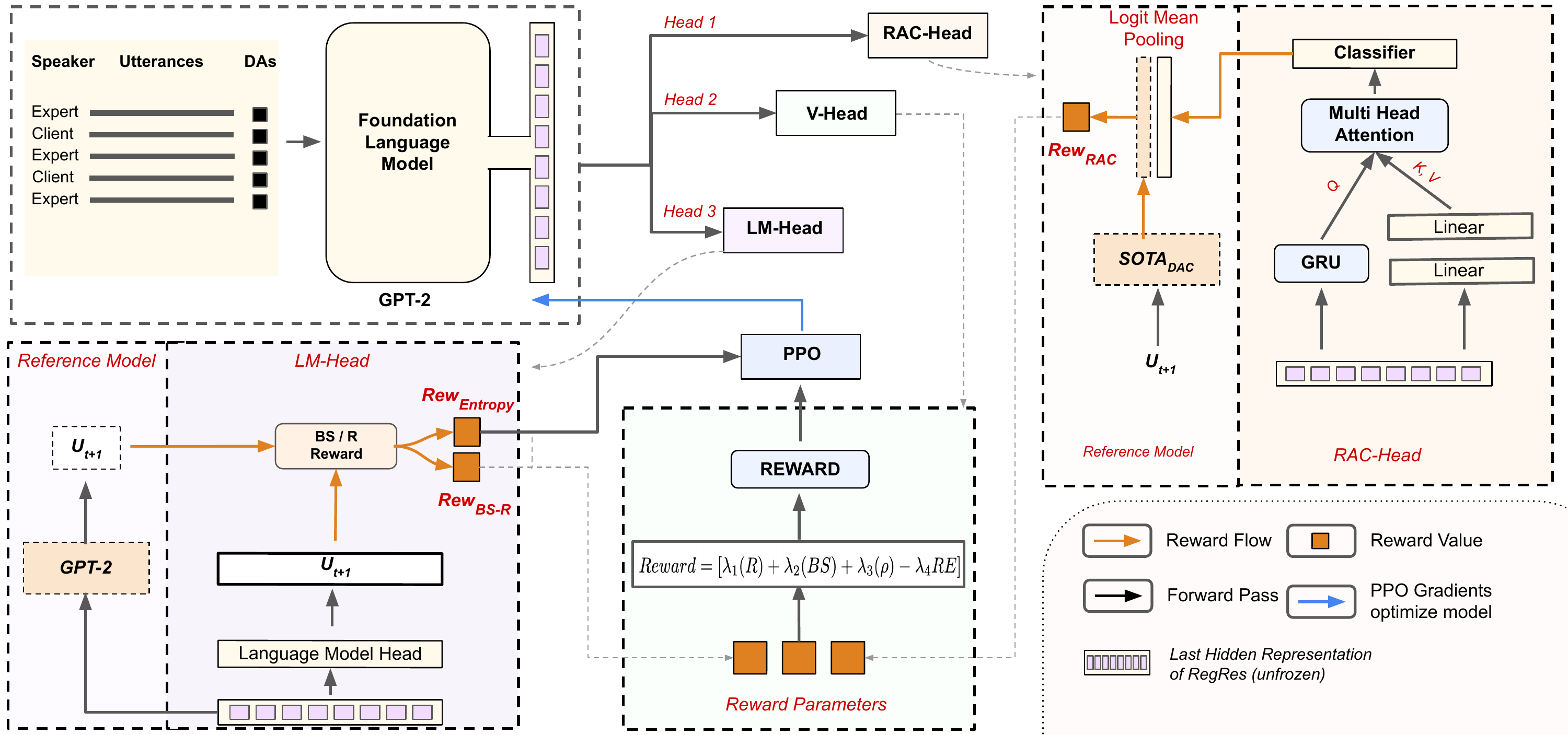}
  % \vspace{-3mm}
  \caption{Architecture of \model. It contains three heads on top of the foundation language model, GPT-2: (a) RAC-Head classifies the response-act trained on context-aware representations, (b) LM-Head generates the response, and (c) V-Head calculates the final reward and initiates Proximal Policy Optimization (PPO).}
  \label{fig:model-architecture}
  % \vspace{-2mm}
\end{figure*}

At the same time, studies by \citet{xu-etal-2019-neural} control responses using meta-words and manually controlled features (\textit{viz.} length of response, specificity, etc.). They defined a meta-word as an organized record. The authors further described the response attributes. This allows them to model the relationship (one-to-many) within task-independent conversations and execute the problem of generating a response in an explainable and controllable manner. %To incorporate meta-words into generation, they proposed a novel goal-tracking memory network that formalizes meta-word expression as a goal in response generation and manages the generation process to achieve the goal with a state memory panel and a state controller.
\citet{NEURIPS2020_e9462095} proposed a simpler architecture which relies on the belief-states generated by the dialogue management module. These belief states are similar to fine-grained intents and slots, exploiting which the authors aimed for the dialogue-generation task.
Further, \citet{khalifa2021a} proposed a  distributional method for handling text generation in a controllable manner by exploiting language models. This method allows point-wise specification of details and distributional constraints on the target language model in one standard framework. Their work is the first effort into this concept while minimizing relative entropy from the earlier proposed language model distribution. They uniquely defined the optimal target distribution as an explicit EBM (Energy-Based Model) representation. Moreover, using those optimal representations, we train a target-controlled autoregressive language model through an adaptive distributional variant of the policy gradient. They conducted experiments on point-wise constraints and showed the advantages of their method over traditional fine-tuning methods. Furthermore, one of the studies on dialogue modeling \cite{DBLP:journals/corr/abs-2008-03391} propose to combine the merits of template-based and corpus-based DRGs by introducing a prototype-based, paraphrasing neural network, called P2-Net, which aims to enhance quality of the responses in terms of both precision and diversity. Instead of generating a response from scratch, they generate system responses by paraphrasing template-based responses. Their approach learns to separate a response into its semantics, context influence, and paraphrasing noise, and to keep the semantics unchanged during paraphrasing. 

\section{Proposed Approach: \model}
In a regular conversation, dialogue-acts of the interlocutors tend to form a pattern. For instance, if person $A$ seeks some clarification from  person $B$, the most probable response from $B$ would be to elucidate the clarification raised by $A$. Therefore, leveraging the above connotation, we propose to utilize the next dialog-act (or response-act) in the response generation task. Formally, we formulate the problem as follows: 
\begin{itemize}[leftmargin=*]
    \item[ ] \em Given a counseling dialogue containing utterances and their corresponding dialogue-acts as $U \in \{u_0,u_1,...,u_{t-1}, u_{t}\}$ and $DA \in \{d_0,d_1,...,d_{t-1}, d_{t}\}$ respectively, where $t$ is the time step, our twofold jointly-learned tasks are -- (a) to predict the response-act $d_{t+1}$ (auxiliary), and (b) to generate a response $u_{t+1}$ in the dialogue abiding by the predicted response-act $d_{t+1}$ (primary). 
\end{itemize}

 To this end, we propose \model, a novel response-act guided reinforced response generation model.  %for mental health counseling. 
The architecture of \model\ is presented in Figure \ref{fig:model-architecture}. \model\ leans on the joint transfer-reinforcement-learning (TRL) paradigm for generating response-acts and responses. Our method of transformer reinforcement learning takes inspiration from an earlier work \cite{DBLP:journals/corr/abs-1909-08593} . Moreover, we train the foundation language model with Proximal Policy Optimization (PPO) \cite{https://doi.org/10.48550/arxiv.1707.06347}.
%Complete TRL with PPO learning takes two parameters as input, namely \textit{current utterance and current dialogue-act} along with the context. 
We define a vocabulary $\Sigma$ and the foundation language model $\theta$ (in our case, GPT2) that defines a probability distribution over sequences of tokens. 

On top of the foundation language model, we place three task-cum-learning specific heads. First, the language model head (LM-Head) is generalized for text generation tasks. Secondly, we introduce a response-act classification head (RAC-Head), an encoder-only model to classify response-acts. At last, we have a value head (V-Head) to compute the reward to send back to the foundation model. 
% \hl{We apply this foundation language model for our task, our input space $X = \Sigma \le m$, data distribution D over X, and output space $Y = \Sigma$, In our case $x \in X$ is utterance and dialogue-act context over a previous window of size $k$ and $y \in Y$ is the generated response and response-act. (Check this sentence.)} 
Next, we train the model jointly to generate responses from LM-Head and predict response-acts from RAC-Head simultaneously. Subsequently, V-Head computes the reward considering the scores of LM-Head and RAC-Head, which in turn is optimized via PPO. We furnish details related to each head and the reward computation in subsequent sections. 

% \subsubsection{RAC-Head.} Dialogue-acts play an essential role in articulating dialogue flow. RAC-Head is a transformer-based encoder-only module on top of the foundation language model that learns to discriminate between the $12$ dialogue-act labels and predict one response-act. In succession, the prediction of response-act allows \model\ to adapt the representations to abide by the predicted response-acts and further generate the desired response. 

\textbf{RAC-Head.} Dialogue-acts play an essential role in articulating dialogue flow. RAC-Head is a transformer-based encoder-only module on top of the foundation language model that learns to predict the future response-act. The head exploits the last hidden representations of the foundation language model. We feed the hidden representations to a GRU to exploit the contextual pattern of the dialog. In parallel, we obtain linear projections of the hidden representation.   
Next, these contextually-rich representations are passed through a multi-head attention module in which we treat the GRU representations as the \textit{query} and the linear projections as the \textit{key} and \textit{value}. Finally, we apply softmax to classify a response-act. 
%In succession, the prediction of response-act allows \model\ to drive the response generation module towards generation of the next utterance with the predicted response-act. 
The prediction calibrates \model\ to adapt the PPO optimization through the RAC-Head's logits, thus allowing LM-Head to generate an appropriate response. 

% Further, we employ GRU \cite{69e088c8129341ac89810907fe6b1bfe} to exploit rich context. Next, these contextually rich representations are given to a multi-head attention module, which in succession is followed by two linear layers. At last, we apply softmax to classify a response-act. The predictions will calibrate \model\ to adapt the PPO optimization allowing LM-Head to create an effective response. In the next section, we will discuss the training of the LM-Head.

\textbf{LM-Head.} We use GPT-2\footnote{https://openai.com/blog/better-language-models/} as our foundation language model. It has been established as one of the preferred models for a variety of generative tasks \cite{https://doi.org/10.48550/arxiv.2112.08718, https://doi.org/10.48550/arxiv.2010.06185, lin-etal-2021-knowledge, 
https://doi.org/10.48550/arxiv.2012.03539}.  

\subsection{Reference Models for Reward Computation}
We aim to augment the response by inheriting adequate semantics and response-acts. To maintain the stability of the reward function, we deploy state-of-the-art reference models to compare the outcomes for both tasks. For the language model head (LM-Head), we employ the pre-trained GPT-2 model as the reference model, whereas, SPARTA \cite{malhotra2021speaker} is used for the response-act head (RAC-Head). Subsequently, we compute ROUGE ($R$), BERTScore ($BS$), and relative entropy ($RE$) between the proposed and the reference language model's outputs. Unlike primitive methods of RL-training with a standard reward function where the model deviates to learn biased features in order to maximize the reward, where for instance, the model may start copying text from reference text to maximize ROUGE, Our approach employs each metric to calculate the reward function and tracks the relative entropy with the performance of reference model in parallel. It ensures that \model's prediction does not deviate significantly and leverages the semantic richness of the pre-trained reference language model. We calculate the relative entropy as follows:
    \begin{equation}
        RE =  \mathbb{E}_{z \sim P_{lm}} [\log P_{lm_p}(z) - \log P_{lm_{ref}} (z)]
    \end{equation}    
where $z$ is sampled from $P_{lm}$, and $lm_{p}$ refers to the proposed language model; whereas $lm_{ref}$ refers to the reference model.  A lower $RE$ score demonstrates better generations; therefore, we employ $RE$ as a direct parameter in the reward computation. 

Similarly, we utilize SPARTA \cite{malhotra2021speaker} to compute the logit values for the predicted dialogue-act and apply mean-pooling for the reward computation.

\textbf{V-Head.} The value head (V-Head) is responsible for accumulating the reward parameters from other heads to yield the reward and subsequently, use it to reinforce the \model. %In this section, we discuss the calculation of reward and possible policy adjustments. 
% We extensively experimented on reward function and find that ROUGE (R), BERTScore (BS), and RAC-Head's logits ($\rho$) collectively, along with relative entropy ($RE$) contribute towards most optimal policy learning.

For our reward function, we use metrics including BERTScore and ROGUE Score. Along with these, we use known response-acts to train a reward model (SPARTA), and then optimize that reward model.

Our proposed reward function accumulates the weighted Rouge score ($R$), BERTScore ($BS$), the relative entropy ($RE$), and the DAC-Head's logit value ($\rho$). The former three components ($R, BS, RE$) offer feedback on the semantic and syntactic richness of the current state, while the last component ($\rho$) guides the model towards the desired response, exhibiting the predicted response-act. We compute the reward as follows:  
\begin{equation}
\label{rewardeqn}
    Reward = [\lambda{_1} (R)+ \lambda{_2}(BS) + \lambda{_3} (\rho) - \lambda{_4}RE]
\end{equation}
where $\lambda_1, \lambda_2, \lambda_3,$ and $\lambda_4$ are hyperparameters and tuned to optimize and maximize the reward. Subsequently, we reinforce the yielded reward to optimize the current state using PPO.

    \subsection{Training and Proximal Policy Optimization}
    Similar to the optimization policy explored by \citet{DBLP:journals/corr/abs-1909-08593} on a general-purpose task, the training of GPT-2 with PPO in \model\ is a three-step process: 
    \begin{itemize}[leftmargin=*]
        \item \textit{\underline{Initiate RAC-Head and LM-Head}:} Given $u_t$, $d_t$ along with the context $\{ <u_{t-k},d_{t-k}>,...,<u_{t-1},d_{t-1}>\}$, where $k$ is the context size, \model\ generates a response-act and the response.
        \item \textit{\underline{Evaluate Outcomes}:} In this step, we calculate the log-probability distribution of logits from the active component (trainable model) of LM-Head and RAC-Head. Subsequently, we obtain the difference in the log probability distribution of reference model which is used to impose penalty and ensure coherency of the outputs.
        % we perform the evaluation on the \model\ to receive outputs from the active \todo{meaningless sentence}. In the next step, we compare the outputs with the reference model to further optimize \model.
        
        \item \textit{\underline{PPO Optimization}:} 
        We choose ROUGE, BERTScore, and Relative entropy to assess the quality of the generated response and the max-logit scores in case of the response-act classification task. We receive the reward score from V-Head (c.f. Equation \ref{rewardeqn}). In order to optimize \model, we first compute the relative entropy from LM-Head. At the same time, LM-Head and RAC-Head yield the remaining reward parameters. At last, V-Head accumulates and computes the reward from each head as per  Equation \ref{rewardeqn}. We perform optimization that subsequently allows \model\ to learn to penalize/reward the foundation language model.
        
    \end{itemize}

\noindent This approach is extended from an earlier work  \cite{DBLP:journals/corr/abs-1909-08593}. The authors initialized a policy $\pi = \theta$, and then fine-tuned $\pi$ to operate on downstream tasks using PPO. If the task is defined by a reward function $(r : X \times Y \rightarrow R)$, then the authors used PPO to optimize the expected reward. In their algorithm, PPO utilizes clipped surrogate objective, and the model maximizes a surrogate objective. 
% as shown below.
% \begin{equation}
%     to write
% \end{equation}
Another study \citet{DBLP:journals/corr/abs-1909-08593} exploited the usage of PPO algorithm to further define the downstream task to optimize the main objective function.
% the following main objective function.
% \begin{equation}
%     to write 2
% \end{equation}
% where epsilon is a hyperparameter. 
The authors opted for minimum of the clipped and unclipped objective. Hence the final objective is lower bound (i.e., a pessimistic bound) on the unclipped objective. With this scheme, we observe that only probability ratio is ignored when it improves the objective.

As a result, we then exploit the above mentioned PPO method to optimize our PPO algorithm \cite{DBLP:journals/corr/abs-1909-08593} with the following equation.
\begin{equation}
 R(x, y) = r(x, y) - \beta\cdot \log\pi(y|x) \cdot \theta(y|x)     
\end{equation}
where $r$ and $\theta$ represent reward function and foundation language model, respectively. In our case, we experiment with a constant as well as dynamic $\beta$ to achieve a favorable value of $RE(\pi, \theta)$. The relative entropy plays the role of an entropy bonus; it  prevents the policy from moving too far from the range where $r$ is valid. We rely on the relative entropy to sync with the fine-tuned reference model's coherent responses.

\begin{table}[!t]
\centering
\caption{Statistics of the HOPE dataset \cite{malhotra2021speaker}. The dyadic counseling conversational dataset contains a total of 12.8k utterances, each associated with one of the 12 dialogue-act labels.}
\label{tab:counts}
\scalebox{1.0}{
\begin{tabular}{l|cccc}
\toprule
HOPE & Train & Validation & Test & Total \\
\cmidrule{1-5}
Dialogue Sessions & 149 & 21 & 43 & 212\\
Client Utterances  & 4668 & 595 & 1119 & 6382\\
Therapist Utterances  & 4751 & 599 & 1122 & 6472\\
\cdashline{1-5}
\#Total Utterances & 9419 & 1194 & 2241 & 12854\\
\bottomrule
\end{tabular}
}
% \vspace{-7mm}
\end{table}

\section{Experiments}
In this section, we first discuss the counseling dataset, HOPE. We then define the baseline systems and evaluation metrics which we use to compare the performance of the proposed model and baselines. 

\subsection{Dataset}
We use HOPE \cite{malhotra2021speaker}, a mental health counseling conversation dataset. %It is compiled from the publicly available counseling conversation dataset. 
It contains $12.8K$ utterances from $212$ dyadic counseling sessions between therapists and clients, publicly available on a video sharing platform. The conversation encompasses diverse demographic groups with distinct mental health discussions. \citet{malhotra2021speaker} extracted transcriptions of the utterances and processed them to remove any noise and/or transcription issues. The collected dialogues are dyadic in nature, i.e., clients and therapists are the only interlocutors. Each utterance in the HOPE dataset is annotated with one of the twelve dialogue-acts -- \textit{information-delivery} (ID), \textit{information-request} (IRQ), \textit{yes/no-question} (YNQ), \textit{clarification-request} (CRQ), \textit{opinion-request} (ORQ), \textit{clarification-delivery} (CD), \textit{positive-answer} (PA), \textit{negative-answer} (NA), \textit{opinion-delivery} (OD), \textit{greeting} (GT), \textit{acknowledgment} (ACK), \textit{general chit-chat} (GC). A detailed statistics of the HOPE dataset is presented in Table \ref{tab:counts}.
Furthermore, we demonstrate the relation between the dialogue-acts of current and next utterances in the HOPE dataset in Appendix (c.f. Figure \ref{fig:chord:plot}). Evidently, the dataset shows a high correlation between certain pairs of dialogue- and response-acts. For instance, an utterance requesting-information (labelled IRQ) is mostly followed by an utterance delivering the information (labelled IRD).

% We can observe that there exists high correlation between certain pairs of acts that frequently intersect together, such as IRQ and ID, YNQ and ID, etc.   

\subsection{Baselines and Evaluation Metrics}
We compare \model's performance with various competitive baselines in the domain of dialogue generation. Moreover, to have a fair comparison with \model, we choose systems which leverage and exploit the dialogue context for the response generation. To the best of our knowledge, no other systems have reinforced the response-act for the dialog generation. We choose the following baselines in this work. 
% Further, Table \ref{tab: results} shows the performance of all the baselines. 
% \begin{itemize}[leftmargin=*]
     \textbf{DialoGPT} \cite{zhang-etal-2020-dialogpt} is a pretrained transformer model dedicated for response-generation task. %We finetune DialoGPT for counseling conversation task to present 
     \textbf{GPT-2} \cite{radford2019language} is a decoder only model trained on a large corpora. A vanilla finetuned version of GPT-2 works well in our use case. %Furthermore, the foundation language model of \model is also GPT-2.
     \textbf{DialogVED}
    \cite{chen-etal-2022-dialogved} introduces continuous latent variables into the encoder-decoder pre-training framework to increase the relevance and diversity of responses.
     \textbf{ProphetNet-Dialog}
    \cite{qi-etal-2021-prophetnet} focuses on pretraining of dialogue specific corpus to generate coherent response.
     \textbf{HRED} \cite{10.5555/3016387.3016435} is based on generative modeling to develop conversational response containing hierarchical encoder-decoder paradigm. It is trained on a large dialogue corpus for the utterance generation task.
     \textbf{HRED with Speaker and Utterance Encoder} \cite{https://doi.org/10.48550/arxiv.1907.05599} adds speaker and utterance level information to the hierarchical encoder-decoder (HRED) setup. It leverages the personalization parameters in a dialogue system. It is also trained on a large dialogue corpus.
     \textbf{VHCR } \cite{park-etal-2018-hierarchical} uses variational hierarchical RNNs for the conversation-only setup. It is trained on a large conversation corpus for the dialogue modeling task. 
% \end{itemize}

\begin{table*}[t]\centering

\renewcommand*{\arraystretch}{1.0}
\caption{Results obtained on the HOPE dataset. We show ROUGE (1, 2, L), BERTScore (BS), and METEOR to assess the performance of the \model. $Rew(x)$ is the reward function, where $x$ is the parameter.}
\label{tab: results}
\resizebox{0.95\textwidth}{!}
{
\begin{tabular}{llcccccccccccc}

\toprule
& \multirow{2}{*}{} &\multicolumn{3}{c}{R1} &\multicolumn{3}{c}{R2} &\multicolumn{3}{c}{RL} &\multirow{2}{*}{BS} &\multirow{2}{*}{METEOR} \\\cmidrule{3-11}
& & P & R & F1 & P & R & F1 & P & R & F1 & & \\ \midrule
\multirow{7}{*}{\rotatebox{90}{Baselines}} & DialoGPT \cite{zhang-etal-2020-dialogpt} &12.34 &40.48 &15.72 &2.92 &11.83 &4.42 &12.23 &38.60 &15.76 &0.7603 &0.2021 \\
& GPT2 \cite{radford2019language} &12.70 & 32.63 & 14.98 & 3.08 & 7.92 & 3.51 & 13.74 & 32.05 & 15.87 & 0.7445 & 0.1754 \\
& DialogVED \cite{chen-etal-2022-dialogved} & 12.48 & 31.74 & 12.8 & 0.98 & 2.45 & 1.22 & 12.45 & 31.11 & 14.46 & 0.7189 & 0.2000 \\
& ProphetNet \cite{qi-etal-2021-prophetnet} &12.15 & 34.29 &14.48 & 3.30 &10.41 &4.17  & 12.24 & 33.12 & 15.18 & 0.6707 & 0.1901\\
& VHCR \cite{park-etal-2018-hierarchical} &11.29 & 21.33 &11.81 & 2.66 &3.49 &3.00  & 10.01 & 19.72 & 10.99 & 0.5953 & 0.1041\\
& HRED \cite{10.5555/3016387.3016435} & 11.52 & 21.51 & 10.72 & 1.89 & 6.42 & 2.92 & 12.12 & 24.36 & 13.56 & 0.6259 & 0.1425 \\
& HRED w/ Sp. Utt. Encoder \cite{https://doi.org/10.48550/arxiv.1907.05599} & 11.77 & 28.63 & 10.08 & 1.29 & 4.19 & 2.06 & 12.25 & 21.27 & 12.72 & 0.6171 & 0.1801\\

\midrule
\multirow{3}{*}{\rotatebox{90}{Ours}} & RagRes w/ DialoGPT &12.41  &43.91 &16.12 &3.70 &13.72 &4.98 &11.92 &41.02 &16.30 &0.7656 &0.2098 \\
& \model\ -- RAC-Head & 12.64 & 41.48 & 15.78 & 3.60 & 11.83 & 4.58 & 12.3 & 38.64 & 15.90 & 0.7628 & 0.2039 \\
% & \rowcolor{blue!20}
% \textbf{Ablations} & & & & & & & & & & & \\
&\bf \model &\textbf{12.82} &\textbf{43.93} &\textbf{16.15} &\textbf{3.77} &\textbf{13.67} &\textbf{4.93} &\textbf{12.51} &\textbf{40.82} &\textbf{16.32} &\textbf{0.7666} &\textbf{0.2103} \\
% \toprule
\midrule
% \textbf{Reward Ablation} & & & & & & & & & & & \\
\multirow{5}{*}{\rotatebox{90}{Ablations}} & $\qquad$ -- $Rew($R$)$ & $11.73$ & $38.82$ & $14.65$& 2.28 &$8.45$&$2.96$& 11.21 &35.76&14.53&0.7561&0.1840 \\
& $\qquad$ -- $Rew($RAC$)$ &12.36 &40.71 &15.43 &3.13 &11.12 &4.06 &11.91 &37.63 &15.40 &0.7609 &0.2000 \\
& $\qquad$ -- $Rew($RAC + R$)$ &11.92 &38.06 &14.70 &2.43 &8.26 &3.11 &11.40 & 34.98 &14.58 &0.7530 &0.1874 \\
& $\qquad$ -- $Rew($R + BS$)$ &12.48 &41.13&15.57 &3.52 &11.85 &4.47&12.22 &38.29 &15.77 &0.7527 &0.2092 \\
& $\qquad$ -- $Rew($RAC + BS$)$ & 12.01 & 40.45 & 15.18 & 2.72 & 9.93 & 3.52 & 11.46 & 37.05 & 14.97 & 0.7577 & 0.1908 \\
% \model\ w/ Rew(R+DAC +BS only) & & & & & & & & & & & \\
\midrule
& $\Delta_{\model-BEST}(\%)$ & \textcolor{blue}{$\uparrow 0.94$} & \textcolor{blue}{$\uparrow 8.5$} & \textcolor{blue}{$\uparrow 2.73$} & \textcolor{blue}{$\uparrow 14.24$} & \textcolor{blue}{$\uparrow 15.50$} & \textcolor{blue}{$\uparrow 11.53$} & \textcolor{red}{$\downarrow 8.90$}& \textcolor{blue}{$\uparrow 5.69$} & \textcolor{blue}{$\uparrow 2.83$} & \textcolor{blue}{$\uparrow 0.82$} & \textcolor{blue}{$\uparrow 4.05$} \\ 

\bottomrule

\end{tabular}}

\end{table*}

% \subsection{}
% We evaluate the responses generated from \model\ on both quantitative and qualitative criteria. \\

% \noindent\textbf{Quantitative Measures.} 
For evaluating the performances of \model\ and other comparative systems, we employ \textbf{ROUGE}, \textbf{METEOR}, and \textbf{BERTScore} as evaluation metrics. We use \textit{py-rouge}\footnote{https://pypi.org/project/py-rouge/}, \textit{nltk-meteor}\footnote{https://www.nltk.org/api/nltk.translate.meteor\_score.html}, and \textit{Hugging Face - BERTScore}\footnote{https://huggingface.co/spaces/evaluate-metric/bertscore} for computing the scores. 

\section{Results and Analysis}
In this section, we discuss the results obtained from the \model\ model and aforementioned baselines. Table \ref{tab: results} summarizes the comparative and ablation results on the HOPE dataset. %The upper half of the table includes comparisons with the baselines, which we discuss in the next subsection, and the lower half of the Table \ref{tab: results} demonstrates the specificity of each module and reward criteria in our proposed approach

\subsection{Performance Comparison}
Our evaluation shows superior performance of \model\ across a majority of the metrics. Evidently, there is a significant increase in the recall of the ROUGE-2 score -- our model receives a ROUGE-2 recall of $13.67$, which is $+15.50\%$ as compared to the second best performer, DialoGPT ($11.83$). At the same time, our model yields $43.93, 40.82, \text{ and } 76.66$ scores of ROUGE-1 recall, ROUGE-L recall, and METEOR, respectively, with an increase of $+3.45, +2.22, \text{and} +0.63$ points as compared to the best baseline, i.e. DialoGPT. 
On the other hand, to evaluate the linguistic properties in the generated utterances, we calculate METEOR on \model's generations. Similar to the earlier cases, \model\ reports an improved BERTScore of $0.2103$, $+4.05\%$ points better than DialoGPT.

\textbf{Ablation on Foundation Model.} Among all baselines, DialoGPT performs the best on average with GPT-2 closely competing with it. However, in our case, we find it suitable to choose GPT-2 as the foundation language model due to the marginally better performance in \model.
We also experiment by swapping it with DialoGPT and report the results in Table \ref{tab: results}. We observe that \model\ with GPT-2 performs better on  recall scores of R1 $(+0.41)$, R2 $(+0.02)$, RL $(+0.2)$, BS $(+0.0139)$ as compared to \model\ with DialoGPT.

\textbf{Model Component Ablation Study.} One of the prime contributions of this work involves the role of the RAC-Head. \model\ relies on the RAC-Head to determine the response-act of the generated utterances. Further, RAC-Head gradually allows the response generation to adapt according to the predicted response-acts during the PPO optimization. We perform ablation on RAC-Head and present the results in Table \ref{tab: results}. We observe that \model\ without RAC-Head scores $41.48$, $11.83$, $38.64$ on ROUGE-1, 2, L (recall scores), respectively. The final model performs relatively better with an increased score ($+2.45, +1.84, +2.18$) on the same metrics as compared to the \model\ without RAC-Head. 

\begin{table*}[]
    \footnotesize
    \renewcommand*{\arraystretch}{1.0}
    \centering
    \caption{An example showing the generated response along with the reference response.}
    \label{tab:output}
    \resizebox{\textwidth}{!}{
    \begin{tabular}{p{30em}|p{30em}|c}
        \toprule
        \bf Context & \bf Next utterance & \bf RAC \\ \midrule 
        \multirow{2}{30em}{\textbf{Therapist:} Oh great, which is a little different. But yeah, that's great. That's a really good goal.} & \multirow{2}{30em}{\textbf{Gold:} Okay, so you're getting A's and B's in really good numbers. So what did you discover about this?} & \multirow{2}{*}{CRQ} \\ 
        % & & \\  
        & & \\ \cmidrule{2-3}
        \multirow{1}{30em}{{\bf Client:} Yeah, thanks.} & \multirow{1}{30em}{\textbf{\model:} Okay. Okay. What classes are you not getting A's and B's in?} & \multirow{1}{*}{CRQ} \\ 
        % & & \\ 
        \cmidrule{2-3} 
        \multirow{2}{30em}{\textbf{Therapist:} It's hard work to get into law school. Are you doing well in school? What kind of grades you get?}  & \multirow{2}{30em}{\textbf{GPT2:} Anyway, I should mention I'm probably going to get through that anyway. That might not be fun.} & \multirow{2}{*}{--} \\  
        % & & \\  
        & & \\ \cmidrule{2-3}
        \multirow{1}{30em}{\textbf{Client:} I'm doing okay. I'm getting mostly A's and B's so I'm pretty good.} & \multirow{1}{30em}{\textbf{DialoGPT:} Are you going to law school?} & \multirow{1}{*}{--} \\ 
        % & \\ 
        \midrule

        \multirow{1}{30em}{\textbf{Client:} My employer just said I wasn't a good fit.} & \multirow{1}{30em}{\textbf{Gold:} My anxiety is really bad.} & \multirow{1}{*}{ID} \\
        % & & \\ 
        \cmidrule{2-3}
        \multirow{2}{30em}{\textbf{Therapist:} wasn't a good fit. So this so these symptoms have caused you some real problems.} & \multirow{2}{30em}{\textbf{\model:} Oh, I get really anxious. just scared that something's going to happen and your anxiety is going to be costly.} & \multirow{2}{*}{ID} \\  
        % & & \\ 
        & & \\ \cmidrule{2-3}
        \multirow{1}{30em}{\textbf{Client:} Yeah.} & \multirow{2}{30em}{\textbf{GPT2:}That's it. Lerner. It’is a honest fact and it is one piece of information about your mind. Lerner} & \multirow{2}{*}{--} \\ 
        % & & \\ 
        \multirow{3}{30em}{\textbf{Therapist:} With employment and how about with your anxiety level?} & & \\ \cmidrule{2-3}
         & \multirow{1}{30em}{\textbf{DialoGPT:} I'm not sure if you're joking or not.} & \multirow{1}{*}{--} \\ 
        % & & \\ 
        \bottomrule
    \end{tabular}}
    \vspace{-3mm}
\end{table*}

\begin{figure}[t]
  \centering
  \includegraphics[width=0.9\columnwidth]{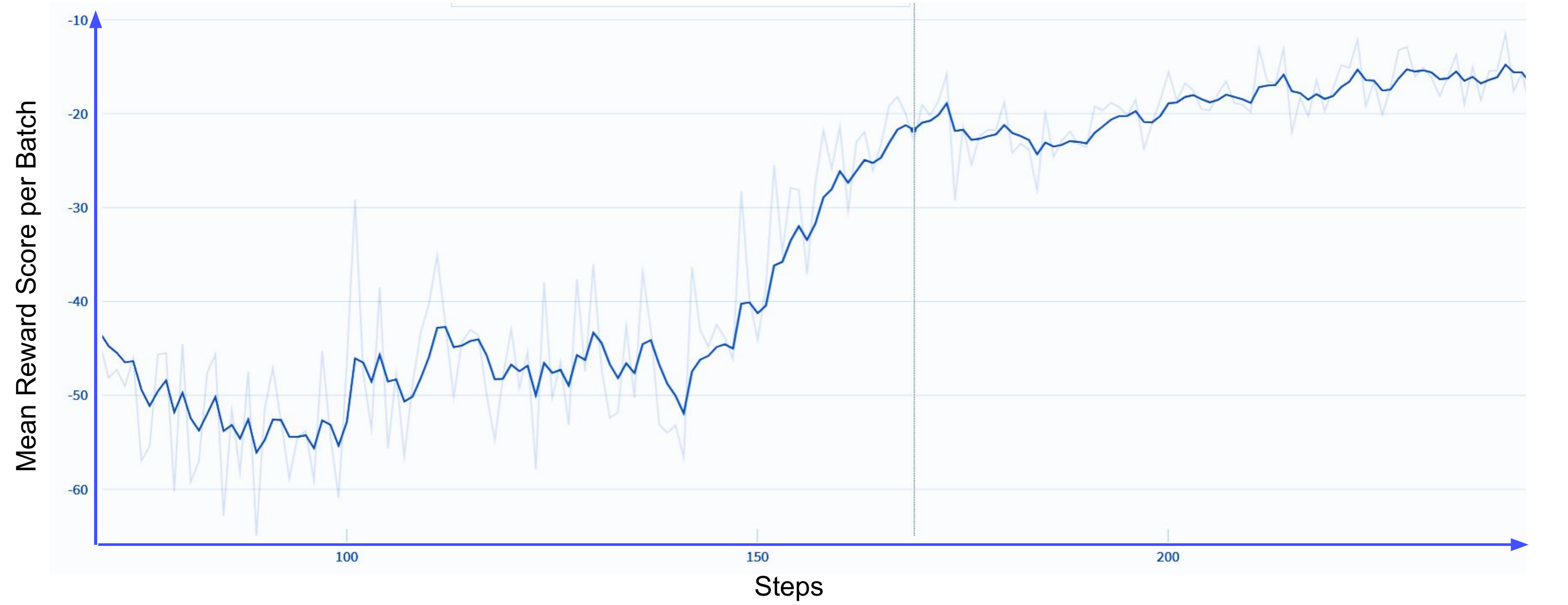}
  \caption{The increment flow of reward in the PPO optimization of \model. 
  }
  \label{fig:rewardplot}
  \vspace{-5mm}
\end{figure}

\textbf{Discussion on Reward Selection and Reward Ablation.}
We meticulously conduct experiments on several hypotheses to design a reward function that optimizes the PPO policy and penalizes the model for every shortcoming. While experimentation, we consider several metric scores as a parameter to the reward function (c.f. Equation \ref{rewardeqn}). However, most of the parametric configurations deteriorate the results. We show various possible ablations on the final set of parameters, i.e., ROUGE, BERTScore, and RAC-Head's logits in the lower half of Table \ref{tab: results}. We observe that a combination of RAC-Head along with BERT Score, ROUGE scores, and relative entropy yields the best result. In addition, using only BERTScore or ROUGE scores in the reward function deteriorates the results significantly. We also observe a decrease in the metrics after detaching RAC-Head. This justifies the contribution of both the heads of \model\ toward the generation task. Moreover, Figure \ref{fig:rewardplot} shows the incremental graph of the reward function during the PPO optimization. Evidently, the plot demonstrates the progressive learning curve considering the mean reward score, and supports our claim that the model improves with the increasing step.

%\vspace{-5mm}
\subsection{Qualitative Analysis}
To further assess the quality of the generated responses of \model, we present a qualitative analysis in Table \ref{tab:output}. In comparison with the reference utterance, we observe that semantics and grammar are well established in the generated response. Additionally, we observe that the response-act of the generated response is in accordance to the intended act. It suggests the effectiveness of RAC-head in reward computation. We also show generated utterances for the two best performing baselines (DialoGPT and GPT). Though the outputs are syntactically correct, they are mostly incoherent with the dialogue context.

%For brevity, we furnish more instances in Appendix. 
% Moreover, the auxiliary task of \model\ is to predict the response-act. We analyze the model's response-act prediction and compare them with the gold labels. The performance further justifies that \model\ continuously adapts itself for jointly learning to predict response-act and generate a response.
% ======================
\begin{table}[!t]
\centering
 % \vspace{-3mm}
\caption{Human evaluation on the responses generated from \model\  when compared to the top two best performing baselines. We observe that the performance of \model\ across all metrics is up to the mark and slightly better than the best-performing dialogue models. }
\label{tab:humaneval}
\scalebox{1.0}{
\begin{tabular}{l|cccc}
\toprule
Model & Relevance & Consistency & Fluency & Coherence \\
\cmidrule{1-5}
DialoGPT  & 2.11 & 2.42 & 2.90 & 2.30\\
GPT2  & 2.70 & 3.00 & 3.01 & 2.44\\
\cdashline{1-5}
\textbf{\model} & \bf 2.85 & \bf 3.05 & \bf 3.05 & \bf 2.95\\
\bottomrule
\end{tabular}
}

 \vspace{-5mm}
\end{table}
% ======================

\textbf{Human Evaluation}
We also perform human evaluation on a subset of model outcomes on linguistic ground. We use four linguistic parameters, namely, \textit{relevance}, \textit{consistency}, \textit{fluency}, and \textit{coherence}, to perform the human evaluation \cite{VANDERLEE2021101151}. We define these parameters as follows: \textbf{Fluency} demonstrates the linguistic quality of the generated responses; \textbf{Coherence} shows the structure and organization of the generated responses; \textbf{Relevance} shows the selection of relevant content in the generated response considering the reference utterance; and \textbf{Consistency} evaluates the factual alignment between the generated response and the source utterance.

In total, we take 50 randomly-selected instances and ask 10 human evaluators to assign a score on a scale of $[1, 5]$ to each of the four parameters, where $5$ represents the best outcome. All human evaluators are linguistic experts, aged between 20 to 35. For comparison, we repeat the exercise for DialoGPT and GPT2 as well. Finally, we compute the average score and report the findings in Table \ref{tab:humaneval}. Our analysis shows that \model's outputs are also qualitatively better than baselines in each dimension.     
% Results are shown in Table \ref{tab:humaneval}. Each parameter is rated on a scale of $1$ to $5$ where a higher score demonstrates better results.

% \noindent\textbf{Qualitative Measures.} Further, we also perform human evaluation on the following four diverse qualitative metrics. %We define the four metrics below. Also, we discuss on human-evaluation in the next section.
% \begin{itemize}

\subsection{Application of \model: Dialogue Generation}
In this section, we present the application of \model\ for generating counseling dialogues. To do so, we adopt two setups: \textbf{a) Natural:} an end-to-end conversation between a client and an agent (\model$_{Therapist}$); and \textbf{b) Synthetic:} an end-to-end conversation between two agents, i.e., \model$_{Therapist}$ and \model$_{Client}$. The first setup is a natural configuration for VMHAs, when deployed at the application stage, it generates therapist utterances to interact with real-time clients having mental health issues. To do so, at every step $i$ of the response generation, we provide actual client inputs and the previously-generated \model's outputs for therapist ($\{i-n, \cdots, i-2, i-1\}$) as the recurring context to \model. On the other hand, the second setup is an analysis configuration to assess the effectiveness of \model\ in handling diverse inputs (e.g., generated by an agent). Moreover, this can also be viewed as a data augmentation technique to generate synthetic dialogues. In this setup, we provide \model's generated outputs for both client and therapist as context.

\begin{figure}[t]
    \centering
    \includegraphics[width=\columnwidth]{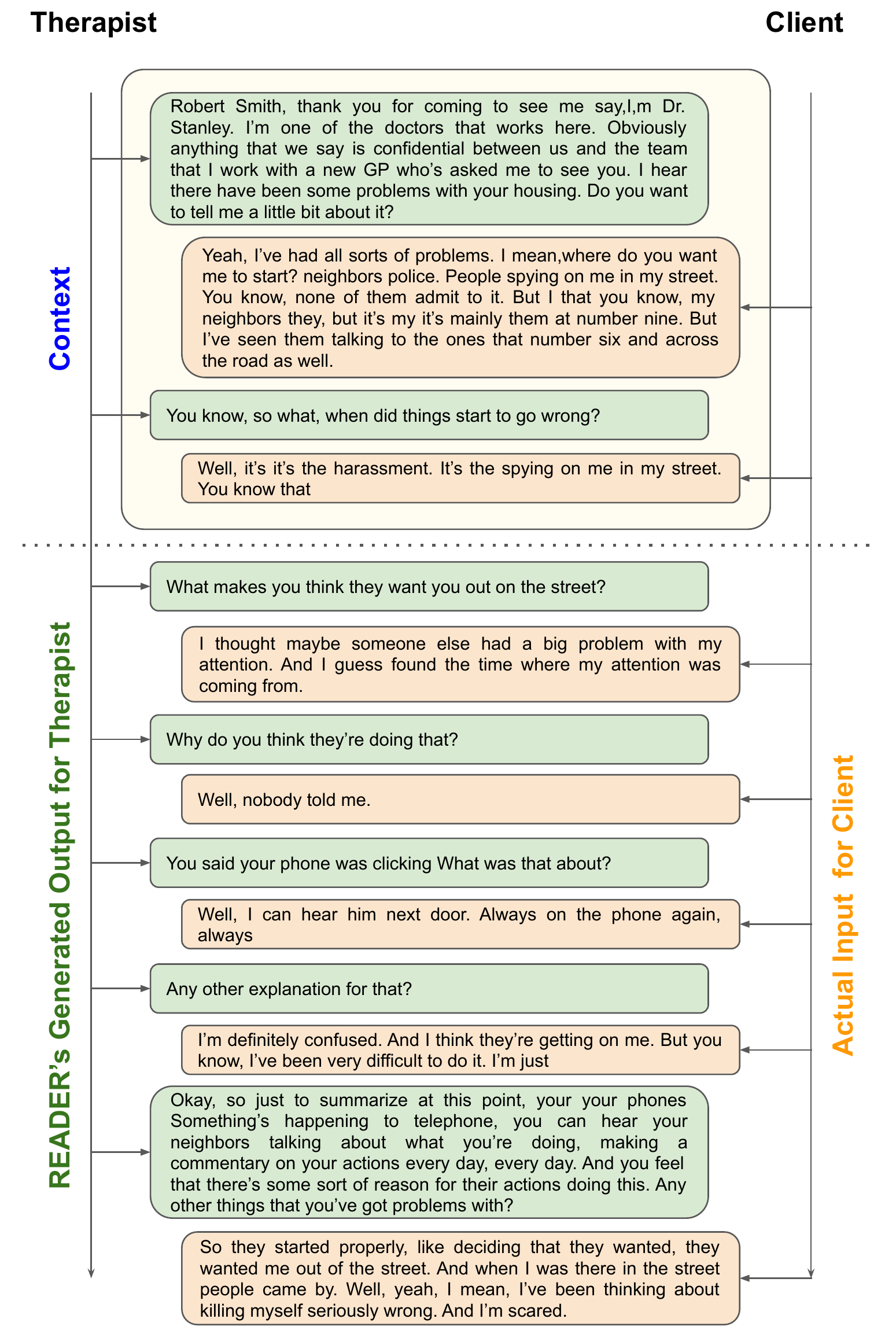}
    \caption{Application of \model\ -- Natural setup. Given a context, at each step, \model\ generates an output for the therapist. This align with the natural configuration of VMHAs, where a client seeks help from a bot or a virtual agent.% complete dialogue conversation generated by the model, \model. The dialogue shows part of utterances from actual conversation and then shows the responses generated considering the previous few utterances as context.
    }
    \label{fig:thepDial}
    \vspace{-5mm}
\end{figure}

\begin{figure}[t]
    \centering
    \includegraphics[width=\columnwidth]{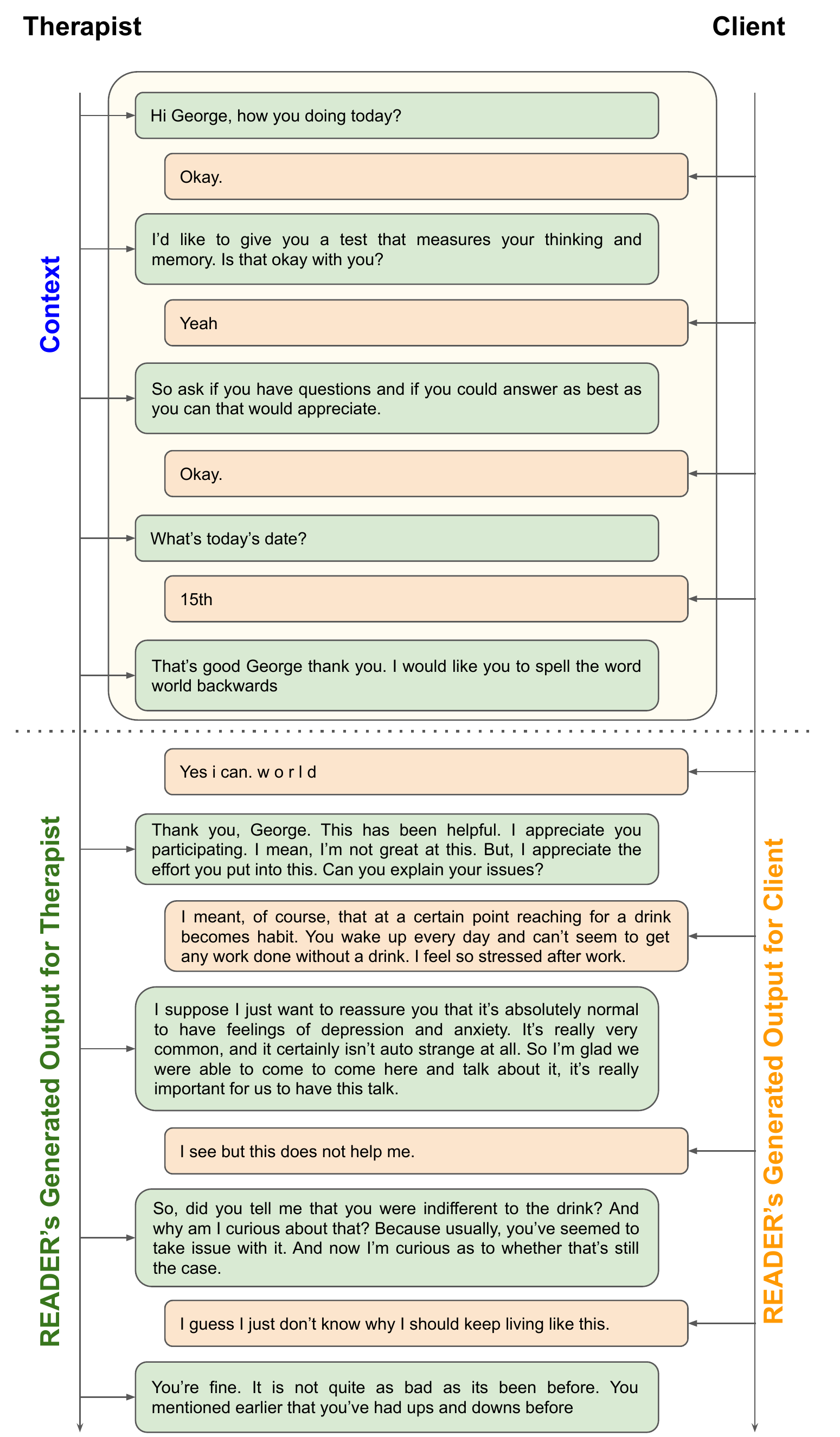}
    \caption{Application of \model\ -- Synthetic setup. Given a context, at each step, \model\ takes turn to generates outputs for the therapist and the client. This setup is an analysis configuration to assess the effectiveness of \model\ in handling diverse inputs (e.g., generated by an agent). %Generation of one complete counseling dialogue turn by turn by \model. We provide \model\ with initial sequence of utterances as an input context and model generates one turn at a time. We repeat this process with generated utterances as new inputs and shown below is a sample of dialogue generated by \model.
    }
    \label{fig:compDial}
    \vspace{-5mm}
\end{figure}

Furthermore, in both setups, we assume that a context is present to instigate the conversation, such that the agent (or \model\ in our case) understands the dynamics of the conversation and starts generating responses that are aligned with the conversation. This approach is similar to existing VMHAs, like WoeBot\footnote{https://woebothealth.com/}, where the agent collects initial information from the client in terms of template-based questions and propels the conversation further with the provided details.     
A snippet of the generated dialogue for the two setups are presented in Figure \ref{fig:thepDial} and Figure \ref{fig:compDial}, respectively. Evidently, we observe that the proposed model is able to comprehend the context of the conversation in both setups and generates aligned responses.

\subsection{Generalizability}
\model\ outperforms several baselines across most of the metrics on the HOPE dataset. Further, to assess the model's generalizability, we extend our experiments and evaluate \model\ on the Switchboard Dialog-act corpus \cite{stolcke-etal-2000-dialogue}. We observe that \model\ improves the performances of two best performing baselines (DialoGPT and GPT2) by $0.1\% - 9.4\%$ in $10$ out of $11$ metrics. In particular, we observe a significant improvement of $9.4\%$ in BERTScore; thus suggesting that \model's outputs are semantically richer than other baselines along with the marginal improvements in textual similarity. %The prime reason for improvement in the BERTScore is highy related to the SWDA dataset's property. 
We argue that in the presence of the information of dialogue-acts, \model\ harnesses the context in an efficient way for generating semantically-richer responses. In conclusion, we posit that \model\ generalized well over other domains as well. Due to the space constraints, we furnish the results along with the baseline's performance on the Switchboard dataset in Appendix (c.f. Table \ref{tab: swda}).

% The results are shown in Table \ref{tab: swda} in the Appendix. We further observe that the \model\ performs better on F1 scores R1, R2, and RL. \model\ also beats competitive baselines -- GPT2 and DialoGPT (c.f Table \ref{tab: swda}). 

\section{Discussion}

\textbf{Societal Impact and Deployment.}  Our work acts as a support to the mental health community and ongoing research by leveraging the advancements in AI-based dialogue systems for counseling. Such advancements in the mental health domain are likely to bring a high social impact. Further, to put this paper’s ideas into practice, we are in active collaboration with a prominent mental health service provider. Collaborators have verified the model's applicability in the real world and agreed to extend \model\ on a bigger corpus and commercialize it. The results of the A/B testing are suppressed due to company's privacy issues.

\textbf{Ethical Considerations and Future Work.}
Considering the severity of the research area, we make sure that at each step, we maintain the privacy of the personal data of clients. In future, we plan to extend our work in the expansion of Virtual Mental Health Assistants (VMHAs) modules and scale the idea of including dialogue components such as empathetic understanding.%, client profiling, etc. %Further, leveraging other languages apart from English is equally important to understand demographic diversities. We plan to incorporate this by the inclusion of minor communities and gender.

\section{Conclusion}
The continuous need to face the shortage in the number of mental health experts is becoming a significant challenge every coming year. With new AI-based therapist-bots coming into the picture, clients receive much support with ease of access. However, one of the critical tasks for such conversational agents is to generate an accurate yet effective response for the clients possessing intended dialogue-act towards the client. To this end, we proposed a novel response-act guided dialogue generation model, \model. We designed a unique reward function that exploits several linguistic properties to train the model using transformer-reinforcement learning (TRL) and further improvised the PPO optimization. We added three heads on top of the foundation language model: RAC-Head, Value-Head, and LM-Head, which collectively curate the reward. We compared the performance of \model\ with several baselines. Our model outperformed several baselines across five metrics: ROUGE (1, 2, \& L), METEOR, and BERTScore. At last, we demonstrated an extensive ablation study and concluded the paper with a discussion on ethical considerations and generalizability.

\begin{acks}
The authors acknowledge the support of ihub-Anubhuti-iiitd Foundation set up under the NM-ICPS scheme of the DST. 
\end{acks}

% bib
\bibliographystyle{ACM-Reference-Format}
\bibliography{sample-base}

\appendix
% \newpage
\clearpage
\section*{Appendix}

\section{Discussion on Optimization}
Our model, \model\ is a result of numerous experiments on the selection of rewards and other hyper-parameters. To scale this model's applicability in the real world, it is equally crucial to understand what worked and what didn't. In this section, we analyze the model's behavior for different configurations. \textbf{Learning Rate (lr): } \model\ behaves unstably with higher learning rate and often causes the reward to collapse significantly. Consequently, we observe that both the convergence rate and the stability of the model start deteriorating.

After fine tuning $lr$ to $2 \times 10^{-7}$, we find an optimal tradeoff point between the convergence of the model and the model's stability. \textbf{Batch-size:} We observe that a large batch size helps with the model's stability for a continuous action space without collapsing the reward. \textbf{Relative Entropy (RE): }We also scaled down the RE value by a factor of 1000 for the reward computation against the standard recommended\footnote{https://github.com/lvwerra/trl} values of \textit{subjective responses} and \textit{highest metrics}.
%We observed joint learning behavior using the Dialogue-Act loss along with Policy loss, Value loss and casual language modelling loss.
% We observe the following observations to the aforementioned experiments: 

\textbf{Reward.} \textbf{a)} Employing the ROUGE-1 score explicitly as a reward is highly prone to the collapse, and as a consequence, the text generation deteriorates through a repetition of similar phrases. \textbf{b)} At the same time, if we use BERTScore only, \model\ becomes unstable and starts generating arbitrarily long responses. %We prevent this by using a static generation method and penalizing utterances longer than a threshold ($\tau$).  
% However, the final result end up under performing compared to a GPT-2 model fine tuned on training data. The generations also don't seem to accurately reflect the joint learning from Dialogue-Act head as well as the dialogue-act generations are inconsistent with predicted response acts. 
\textbf{c)} Using an external reference classifier (i.e., SPARTA) to supplement the reward shows significant improvement in generation quality and benchmark metrics. It ensures that the generated responses are consistent with the response-acts gold labels.

\begin{figure}[H]
    \centering
    \includegraphics[width=0.8\columnwidth]{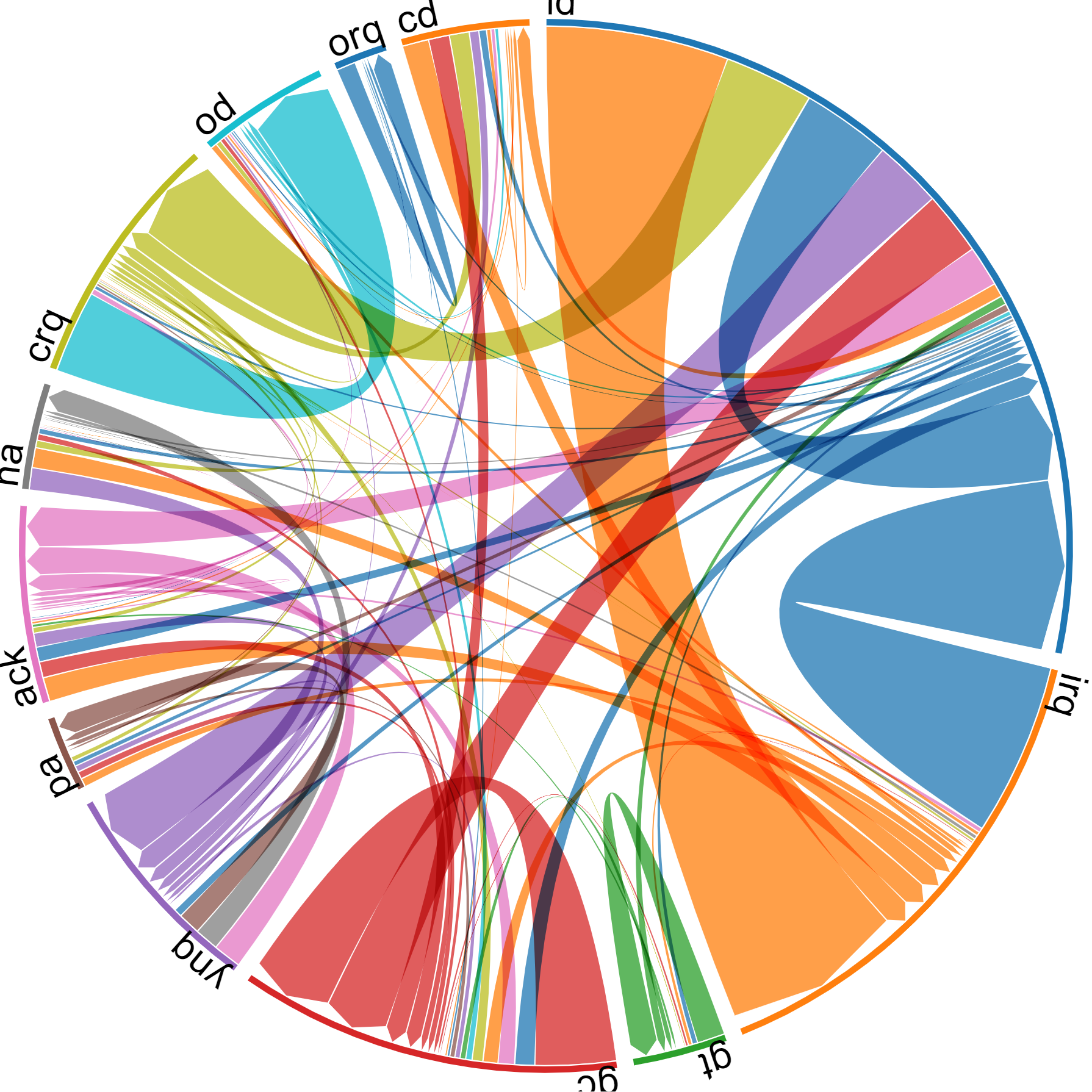}
    \caption{Inspired by the work of \citet{malhotra2021speaker}, we take their proposed relationship among dialogue-act classes. Here, the directed path $U_t^x \rightarrow U_{t+1}^y$ demonstrates the current dialogue-act and response-act pair. \textit{Note: We use this diagram directly from their work}.}
    \label{fig:chord:plot}
    % \vspace{-4mm}
\end{figure}

\begin{table}[!t]
\centering
\caption{ The table demonstrate the results of RAC-Head on the response-act classification task. The comparision of the head is three-fold: a) reference model (SOTA) vs gold dialogue-act labels, b) RAC-Head's prediction vs gold labels, and c) RAC-Head's prediction vs reference model (SOTA). We show the accuracy along with weighted precision, recall, and F1 scores. }
\label{tab:rac_head_per}
 % \vspace{-3mm}
\scalebox{0.9}{
\begin{tabular}{l|cccc}
\toprule
Model & Precision & Recall & F1 & Accuracy \\
\cmidrule{1-5}
SOTA vs Gold Labels  & 0.69 & 0.55 & 0.52 & 0.55\\
RAC-Head vs Gold Labels  & 0.49 & 0.49 & 0.42 & 0.49\\
RAC-Head vs SOTA  & 0.50 & 0.45 & 0.41 & 0.45\\
\bottomrule
\end{tabular}
}
% \vspace{-5mm}
\end{table}

\section{Experimental Setup}
We perform numerous experiments using various combinations of the autoregressive language modeling loss, the dialogue-act loss, the value loss, and the policy loss. Moreover, we conduct extensive hyper-parameter tuning to correctly optimize the PPO trainer and scaling of the relative entropy reward. Further, we extensively experimented with the reward function and observe that Rouge, BERTScore, and RAC-Head's logits along with the relative entropy ($RE$) contribute towards most optimal policy learning.

We perform all experiments on an Nvidia A6000 GPU. We tune our hyperparameters to find the optimal configurations. We utilize the learning rate of $2x10^{-6}$, batch size of $128$, which we run for $4$ PPO-epochs. We use the Adam optimizer and train the reference \model\ for 50 epochs. We also perform hyperparameter tuning on values of $\lambda$ in Equation \ref{rewardeqn} and observe that \model\ works best with $\lambda_1=0.5; \lambda_2=0.15; \lambda_3=0.15; \lambda_4=0.2$.

\begin{table*}[!t]\centering
\footnotesize
% \scriptsize
% \renewcommand*{\arraystretch}{1.0}
\caption{Results obtained on the Switchboard Dialogue-act dataset. We show Rouge (1, 2, L), BERTScore (BS), and Meteor to assess the generalizability of the \model\ on datasets similar to HOPE.}
\label{tab: swda}
\resizebox{\textwidth}{!}
{
\begin{tabular}{lcccccccccccc}\toprule
\multirow{2}{*}{} &\multicolumn{3}{c}{R1} &\multicolumn{3}{c}{R2} &\multicolumn{3}{c}{RL} &\multirow{2}{*}{BertScore} &\multirow{2}{*}{Meteor} \\\cmidrule{2-10}
& P & R & F1 & P & R & F1 & P & R & F1 & & \\\midrule
DialoGPT &21.64 &27.16 &23.87 &11.41 &14.43 &12.47 &19.53 &23.78 &21.46 &0.6615 &0.1836 \\
GPT2 & 22.25 &27.95 &24.14 &11.89 &14.49 &12.75 &20.12 &24.76 &21.80 &0.6608 &0.1822 \\
\midrule
\rowcolor{blue!20}
\bf \model & 22.32 &27.92 &24.18 &11.97 &14.49 &12.79 &20.20 &24.76 &21.84 & 0.7295 & 0.1850\\

\midrule
$\Delta_{\model-BEST}(\%)$ & \textcolor{blue}{$\uparrow 0.3$} & \textcolor{red}{$\downarrow 0.1$} & \textcolor{blue}{$\uparrow 0.1$} & \textcolor{blue}{$\uparrow 0.6$} & \textcolor{blue}{$\uparrow 0.0$} & \textcolor{blue}{$\uparrow 0.3$} & \textcolor{blue}{$\uparrow 0.3$}& \textcolor{blue}{$\uparrow 0.0$} & \textcolor{blue}{$\uparrow 0.1$} & \textcolor{blue}{$\uparrow 9.4$} & \textcolor{blue}{$\uparrow 1.5$}  \\

\bottomrule
\end{tabular}}

\end{table*}
\section{Performance of RAC-Head}
We deploy RAC-Head to perform the task of response-act classification on the HOPE dataset. RAC-Head and LM-Head jointly learns to optimize the \model. Further, to analyze the performance of RAC-Head, we present the results in Table  \ref{tab:rac_head_per}.

\section{Analysis}
We present a detailed analysis of responses generated by \model\ in Table $7$. We see the model is able to correctly incorporate the context. Further observations shows that the quality of the generations is up to the mark and \model\ is capable to carry out full-fledged counseling. However, considering the severity of the matter, we never intend to eliminate the human in the loop.

\section{Generalizability}
We discuss the generalizability of our proposed model, \model\ on the Switchboard Dialogue-act Corpus for the response-generation task. We present the results in the Table \ref{tab: swda}. The results show that the performance of the model is better on majority of the metrics. 

\section{Reproducibility Checklist}
We upload our code in a zip file to reproduce the results of \model. In this section we show the directory structure. The root directory consist of modified trl sub directories consisting of model file named $gpt2.py$, ppo trainer file named $ppo.py$ along with additional helper functions in $core.py$ file. The training program is stored in $Train.py$ file and the latest saved checkpoint can be located in $model\_train\_final.pt$. Further details can be found in an enclosed README.md file.

\end{document}